\documentclass{article}

\usepackage[final]{Styles/neurips_2024}

\usepackage{graphicx}

\usepackage[utf8]{inputenc} 
\usepackage[T1]{fontenc}    
\usepackage{hyperref}       
\usepackage{url}            
\usepackage{booktabs}       
\usepackage{amsfonts}       
\usepackage{nicefrac}       
\usepackage{microtype}      
\usepackage{xcolor}         
\usepackage{tcolorbox}
\usepackage{amsmath} 

\definecolor{myblue}{RGB}{64,64,64}
\definecolor{mygray}{gray}{0.9}

\tcbset{
  myboxstyle/.style={
    rounded corners,
    boxrule=1mm,
    colback=mygray,
    colframe=myblue,
    fonttitle=\bfseries,
    coltitle=white,
    boxsep=5mm,
    left=5mm,
    right=5mm,
    top=5mm,
    bottom=5mm,
  }
}

\title{SAC-GLAM: Improving Online RL for LLM agents with Soft Actor-Critic and Hindsight Relabeling}

\author{
  Loris Gaven \\
  Inria (Flowers)\\
  University of Bordeaux, France  \\
  \texttt{loris.gaven@inria.fr} \\
  \And
  Clément Romac \\
  Inria (Flowers) \\ 
  University of Bordeaux, France \\
  Hugging Face  \\
  \And
  Thomas Carta \\
  Inria (Flowers)\\
  University of Bordeaux, France  \\
  \And
  Sylvain Lamprier \\
  Univ Angers, LERIA, \\ SFR MATHSTIC, F-49000 Angers, France\\
  \And
  Olivier Sigaud \\
  Sorbonne Université, ISIR, Paris, France \\
  \And
  Pierre-Yves Oudeyer \\
  Inria (Flowers) \\
  University of Bordeaux, France \\
}

\begin{document}

\maketitle

\begin{abstract}
  The past years have seen Large Language Models (LLMs) strive not only as generative models but also as agents solving textual sequential decision-making tasks. When facing complex environments where their zero-shot abilities are insufficient, recent work showed online Reinforcement Learning (RL) could be used for the LLM agent to discover and learn efficient strategies interactively. However, most prior work sticks to on-policy algorithms, which greatly reduces the scope of methods such agents could use for both exploration and exploitation, such as experience replay and hindsight relabeling. Yet, such methods may be key for LLM learning agents, and in particular when designing autonomous intrinsically motivated agents sampling and pursuing their own goals (i.e. autotelic agents). This paper presents and studies an adaptation of Soft Actor-Critic and hindsight relabeling to LLM agents. Our method not only paves the path towards autotelic LLM agents that learn online but can also outperform on-policy methods in more classic multi-goal RL environments.  
\end{abstract}

\section{Introduction}
In recent years, LLMs have demonstrated impressive capabilities in solving various tasks, from following instructions to complex reasoning and real-world interactions \citep{ahn_as_2022,li_pre-trained_2022,wang_describe_2023,wang_jarvis-1_2023}. Despite these successes, challenges remain in applying LLMs as agents in sequential decision-making tasks, mainly due to misalignment when understanding environmental states and action spaces. In a recent line of work \citep{carta2023grounding,wen_entropy-regularized_2024,putta_agent_2024,wen_reinforcing_2024} the use of online RL has been explored to interactively align LLM agents with complex environments. Following these promising results, one could wonder whether such agents could learn in a more autotelic way where goals are not directly provided by the environment.

A natural next step would be giving these LLM agents more autonomy, allowing them to freely explore environments and set their own goals \citep{colas_autotelic_2021}. When designing such agents, prior work notably showed how language can be leveraged to imagine new goals and shape the agent's curriculum \citep{colas_vygotskian_2022, colas2020language}. Recent studies successfully showed LLMs could generate goals \citep{wang_voyager_2023,zhang_omni_2024} and reward functions \citep{fan_minedojo_2022,du_guiding_2023} for an autonomous agent, leveraging their world-modeling capabilities and commonsense knowledge to drive the agent towards interesting goals.
While mixing prior work using LLMs as goal generators and LLMs as RL agents may appear straightforward, several challenges remain in obtaining autotelic LLM agents.

Among them, autotelic architectures usually involve experience replay and hindsight relabelling to best utilize all trajectories, even unsuccessful ones, which can be predominant when agents set their own goals. However, most current approaches to fine-tune LLM agents with RL rely on on-policy algorithms that cannot handle experience replay and trajectory relabeling mechanisms. While a handful of attempts \cite{wen_entropy-regularized_2024,putta_agent_2024} to use off-policy RL on LLMs exist, they consider token-level actions (better suited for text generation tasks) and complex architecture balancing token-level actions and environment-level actions (usually sequences of tokens). As a step towards autotelic LLM agents, this work introduces a version of Soft Actor-Critic (SAC) \citep{haarnoja_soft_2018} explicitly designed for LLMs as RL agents combining the simplicity of prior on-policy methods focusing on environment-level actions (e.g. GLAM \citep{carta2023grounding}) and the advantages of off-policy learning. We then extend it with Hindsight Experience Replay (HER) \citep{andrychowicz2017hindsight} and show our new method (SAC-GLAM) outperforms GLAM (called PPO-GLAM below) in a classic multi-goal environments while showcasing better sample efficiency while staying on par in time efficiency in spite of the stability and efficiency challenges of using an actor-critic approach with a pre-trained LLM actor and a randomly initialized critic.




\section{SAC-GLAM}
We consider a textual RL setting where, given a language vocabulary $\mathcal{V}$, the environment returns an observation $ o \in \mathcal{V}^N $ and a reward $ r \in \mathbb{R} $ after an action $ a \in \mathcal{A} \subseteq \mathcal{V}^M $ (i.e., actions are sequences of tokens). The task or goal description $ g \in \mathcal{G} \subseteq \mathcal{V}^K $ conditions the reward. This environment can be modeled as a goal-augmented Partially Observable Markov Decision Process (POMDP) $ \mathcal{M} = (\mathcal{S}, \mathcal{V}, \mathcal{A}, \mathcal{T}, \mathcal{R}, \mathcal{G}, \mathcal{O}, \gamma) $, where $\mathcal{S}$ is the state space, $\mathcal{A}$ is the action space, $\mathcal{G}$ is the goal space, $ \mathcal{T}: \mathcal{S} \times \mathcal{A} \to \mathcal{S} $ is the transition function, $ \mathcal{R}: \mathcal{S} \times \mathcal{A} \times \mathcal{G} \to \mathbb{R} $ is the goal-conditioned reward function, $ \mathcal{O}: \mathcal{S} \to \mathcal{V}^N $ is the observation function that maps a state to a textual description, and $ \gamma $ is the discount factor.

In such settings, PPO shined for its efficiency and simplicity when finetuning stochastic pre-trained policies such as LLM agents \citep{carta2023grounding,wen_reinforcing_2024}. However, since it is an on-policy method, it cannot leverage HER. We propose in sections below an off-policy alternative that keeps GLAM's simplicity on how the LLM is used as a stochastic policy over discrete environment-level actions (i.e. sequences of tokens).

\begin{figure}
    \centering
    \includegraphics[width=1.0\linewidth]{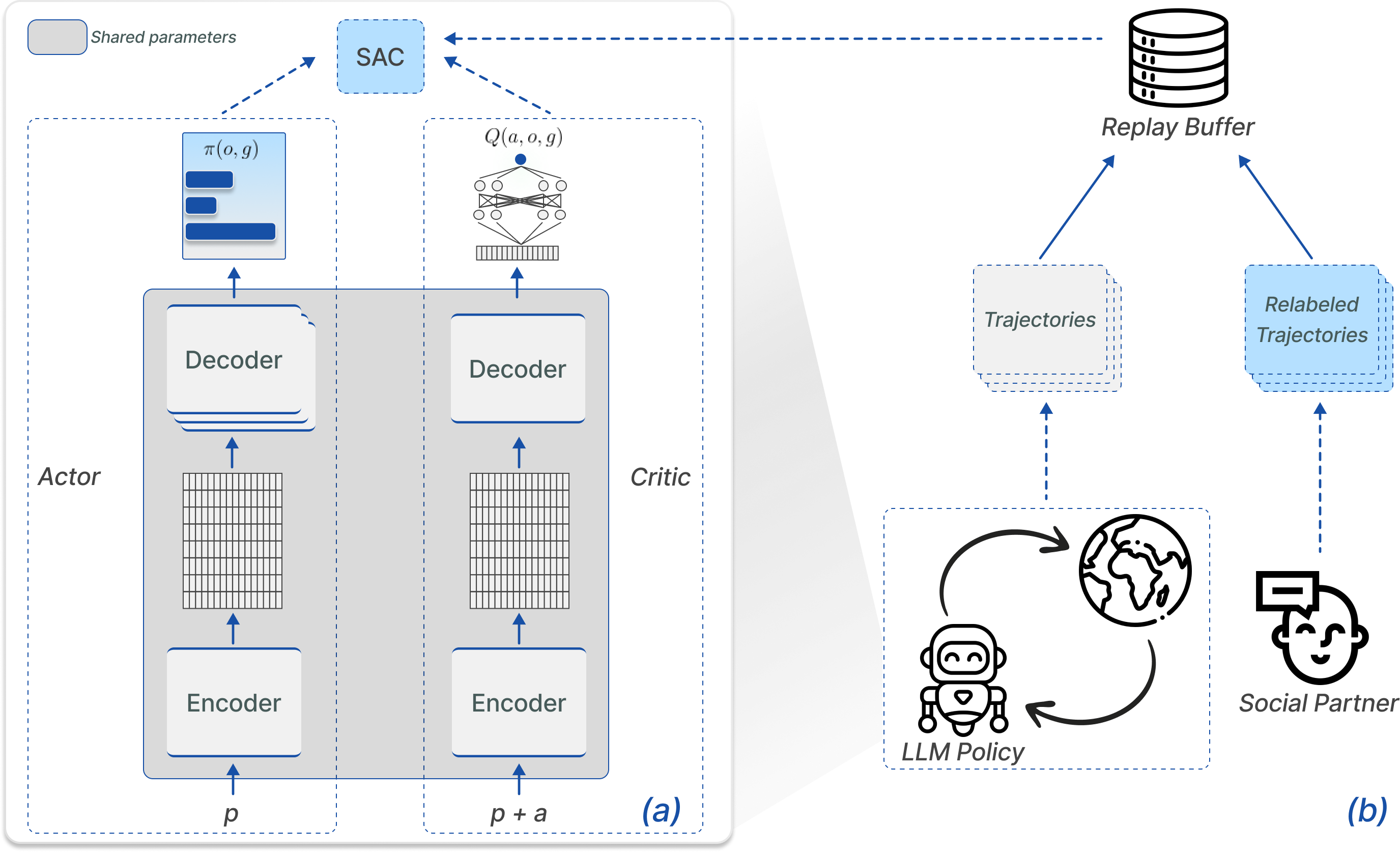}
    \caption{\textbf{The SAC-GLAM method.} (a) depicts the agent's architecture when an encoder-decoder LLM is used: the actor computes an action probability as the probability computed by the LLM of the action's tokens to follow the observation and goal concatenated in the prompt $p$, while the critic computes the Q-value for each action $a$ and the prompt $p$ with an MLP attached to the decoder's last hidden state. (b) illustrates the agent-environment interaction, where trajectories are generated and added to the replay buffer. We used an environment where a social partner relabels these trajectories with hindsight goals.}
    \label{fig:enter-label}
\end{figure}

\subsection{Soft Actor Critic with an LLM actor} \label{subsec:sac_llm}
We adapt the discrete version of SAC \citep{christodoulou2019soft} to our stochastic LLM-based policy. We first follow GLAM's approach to obtaining a stochastic policy with an LLM by computing the probability of each action $ a_i \in \mathcal{A} $ from our environment as the probability of its token sequence $ a_i = \{ w_1, w_2, ..., w_{|a_i|} \} $ (with $w_i \in \mathcal{V} $) to follow a prompt containing both $o$ and $g$:

\begin{equation}
    \mathbb{LP}_{LLM}(a_i|o,g) = \sum_{j=0}^{|a_i|} \log \mathbb{P}_{LLM}(w_j|o,g,w_{<j}) \,.
\end{equation}

Subsequently, a softmax function is applied to generate a probability distribution over all possible actions and obtain our stochastic policy:

\begin{equation} \label{eqn:actor}
    \pi(a_i | o, g) = \frac{e^{\mathbb{LP}_{LLM}(a_i|o,g)}}{\sum_{a_j \in \mathcal{A}} e^{\mathbb{LP}_{LLM}(a_j|o,g)}} \,.
\end{equation} 

Then, we follow SAC and simultaneously learn both this policy and a Q-function, which we implement as an MLP added on top of the final decoder block of our LLM.

\subsection{Pre-trained policy and randomly initialized critic}
\label{sec:AC_stable}
When running SAC with a pre-trained LLM as the policy, several unique challenges arise. In SAC, the actor's updates rely on feedback from the critic. While poor estimates from a randomly initialized critic may not significantly impact a randomly initialized policy, our situation is different. We aim to protect the pre-trained LLM actor from harmful updates that could degrade its performance. One potential solution is to introduce a warm-up period where only the critic is trained. 

To make this warmup period as short as possible, one must seek for faster critic convergence. First, one could speed up its learning not only backpropagating gradients through the MLP but also through the LLM's weights, while potentially impairing the actor's pre-trained capabilities. Second, while a natural choice would be to have an MLP with $|\mathcal{A}|$ outputs that take as input the representation produced by the LLM for the observation and goal in the prompt, we also investigated leveraging the LLM's capabilities by rather using an MLP with a single output with the LLM encoding the concatenation of the prompt and an action. This requires $|\mathcal{A}|$ forward passes through the LLM and MLP to obtain all the actions' Q-value but may also ease the MLP's task leading to faster convergence.  

Finally, the critic's sample efficiency can also be improved after the warmup period. One way to achieve this is through the use of n-step returns with a target expressed as: 



\begin{equation}
y_t = \sum_{i=0}^{n-1} \gamma^i r_{t+i} + \gamma^n \mathbb{E}_{a_{t+n} \sim \pi(\cdot \mid o_{t+n}), g} \left[ Q(s_{t+n}, a_{t+n}) - \alpha \log \pi(a_{t+n} \mid o_{t+n}, g) \right] \,.
\end{equation}
In the standard SAC algorithm, a single-step return is used, which corresponds to setting \( n = 1 \). This approach reduces the reliance on bootstrapping by accumulating rewards over multiple steps before bootstrapping, often leading to faster critic convergence. However, n-step returns introduce the risk of older transitions becoming highly off-policy, as intermediate rewards and actions depend on the policy in place when the transitions were collected. To mitigate this issue, we focus on minimizing the age of the oldest transitions (i.e., the number of policy updates since the transition was recorded). One effective way to achieve this, without altering the update-to-data ratio, is by reducing the size of the replay buffer.

\subsection{Hindsight Experience Replay}
We augment our SAC-GLAM agent with HER, enabling the algorithm to learn from failed attempts by relabeling portions of the failed trajectories with accidentally reached goals. At the end of an episode, the trajectory with its initial goal and the relabeled sections of the trajectory are added to the replay buffer. When training, we randomly sample a batch transitions from this buffer such that the batch contains $50\%$ of relabeled transitions.

\section{Experiments}
We evaluated our method in the Playground-text environment, a text-based adaptation (goals, observations, and actions are represented as text) of the original Playground environment \citep{colas2020language}. We selected this environment because it features a social partner describing goals reached during a trajectory. In this environment, the agent receives a textual goal and must interact with different objects (plants, animals, food and water) to complete the task. It features a sparse reward context where the agent receives $1$ if the task is completed and $0$ otherwise. In the original Playground environment, the agent moves along the $x$ and $y$ axes to reach objects. In the text-based version, we simplified this by introducing actions like "\textit{Go to \{object\}}", which directly move the agent to the selected object. The environment features two types of tasks: "\textit{Grasp \{object\}}" and "\textit{Grow \{object\}}". Animals can be grown using either water or food, while plants can only be grown using water. We also introduced sequential tasks, where the agent must complete two tasks in the correct order, significantly expanding the task space to $8\text{,}470$ tasks. A detailed description of the environment is given in Appendix~\ref{sec:env}.

\begin{figure}[htbp]
    \centering
    \includegraphics[width=\linewidth]{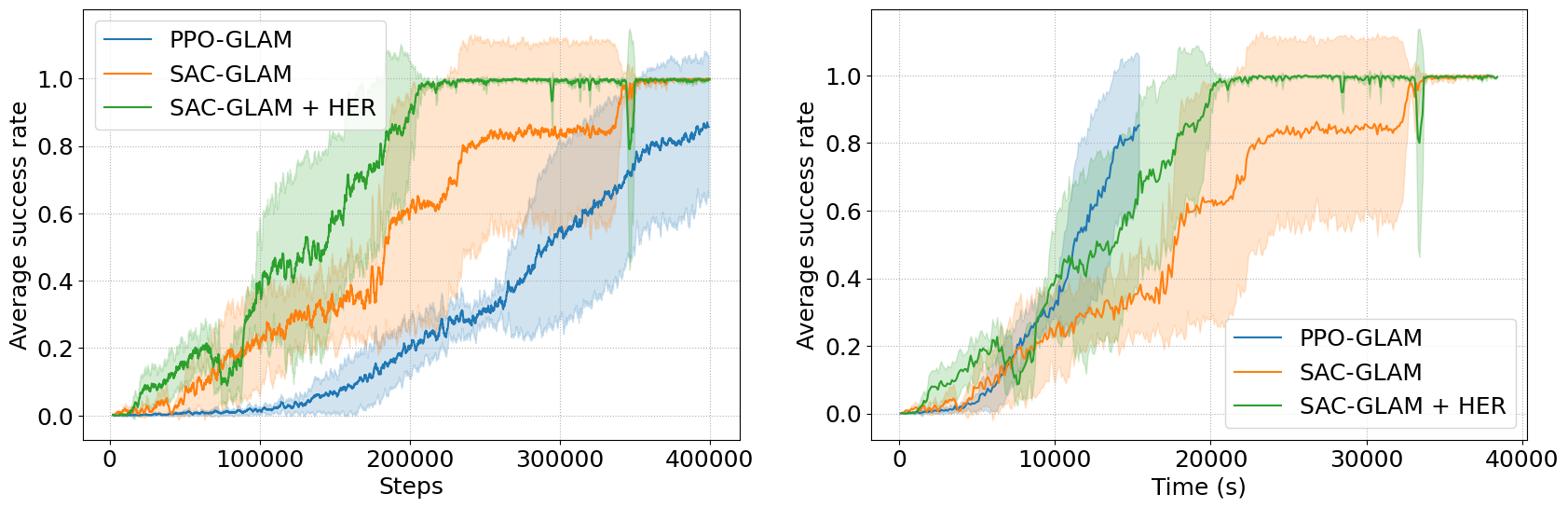}
    \caption{\textbf{Performance comparison of SAC-GLAM and PPO-GLAM in the Playground-Text environment.} We show the average success rate as a function of the number of steps (left) and the average success rate over time in seconds (right). The mean and standard deviation are calculated across 4 seeds.}
    \label{fig:sac_ppo}
\end{figure}

We compared SAC-GLAM with and without HER to PPO-GLAM using Flan-T5-base \citep{chung2024scaling} as the pre-trained actor for both approaches.  Our results in Figure \ref{fig:sac_ppo} indicate that for a fixed budget of $400$K steps, SAC-GLAM alone outperforms PPO-GLAM in sample efficiency while being slower. When equipped with HER, SAC-GLAM becomes even more sample efficient and achieves comparable performance in terms of training time.

\subsection{Ablations}
We study in Appendix~\ref{sec:archi} different possible architectures for the critic as proposed in Section \ref{sec:AC_stable}. We show that sharing the policy's weights and using an MLP with a single output leads to better sample efficiency and greater stability. We hypothesize that the increased stability is due to the latent representation provided to the MLP, which directly incorporates information about both the prompt and the action.

We also study the impact of the update frequency and number of epochs (Appendix~\ref{sec:update_freq}), replay buffer sampling strategies (Appendix~\ref{sec:sampling}) and n-step returns (Appendix~\ref{sec:n_step}) on the critic's convergence speed. We found that doing frequent updates (every 32 steps) with multiple epochs (2), using $n=3$ and balancing transitions relabeled by HER in batches (50\%) was leading to the best performance. 

\section{Discussion}

In this paper, we introduce SAC-GLAM, an off-policy RL approach adapting SAC and HER to LLMs as RL agents. We notably highlight how Actor-Critic methods applied to pre-trained policies such as LLMs introduce challenges in balancing the policy and critic convergence and showed using a warmup period, weights sharing, single Q-value head and n-step returns make SAC-GLAM obtain comparable time efficiency to PPO-GLAM while surpassing it in sample efficiency. Finally, in addition to outperforming PPO-GLAM in a classic multi-goal RL setting, SAC-GLAM also paves the path toward autotelic LLM agents where using HER and off-policy RL is key.

\begin{ack}
Experiments presented in this paper were carried out using the HPC resources of IDRIS under the allocation 2023-[A0151011996] made by GENCI.
\end{ack}

\bibliography{main}

\newpage
\appendix

\section{Playground-Text}
\label{sec:env}

The Playground-Text environment is a text-based environment where goals, observations, and actions are all represented in natural language. The environment consists of a scene with N=6 objects. The agent can move toward these objects using the action "\textit{Go to \{object\}}", and can interact with them using "\textit{Grasp}" to pick up objects and "\textit{Release}" to put them down. There are four types of objects: furniture, supplies, plants, and animals. The "supplies" category contains two items: food and water. When the agent brings water to a plant, it grows; animals can grow with either water or food. The colors of the objects are irrelevant and serve only as distractions for the agent. Observations from the environment fully describe its state, as shown in Figure \ref{fig:prompt}.

\begin{figure}[htbp]
\centering
\begin{tcolorbox}[myboxstyle]
Goal: Grow green mouse then grow blue rose\\
You see: green water, green mouse, blue rose, red rose, blue lion, red food\\
You are on: nothing\\
You are holding: nothing\\
You have grown: nothing\\
Action: 
\end{tcolorbox}
\caption{\textbf{An observation in the Playground-Text environment.} All necessary information is provided in the observation, making the environment fully observable.}
\label{fig:prompt}
\end{figure}

\begin{figure}
    \centering
    \includegraphics[width=0.7\linewidth]{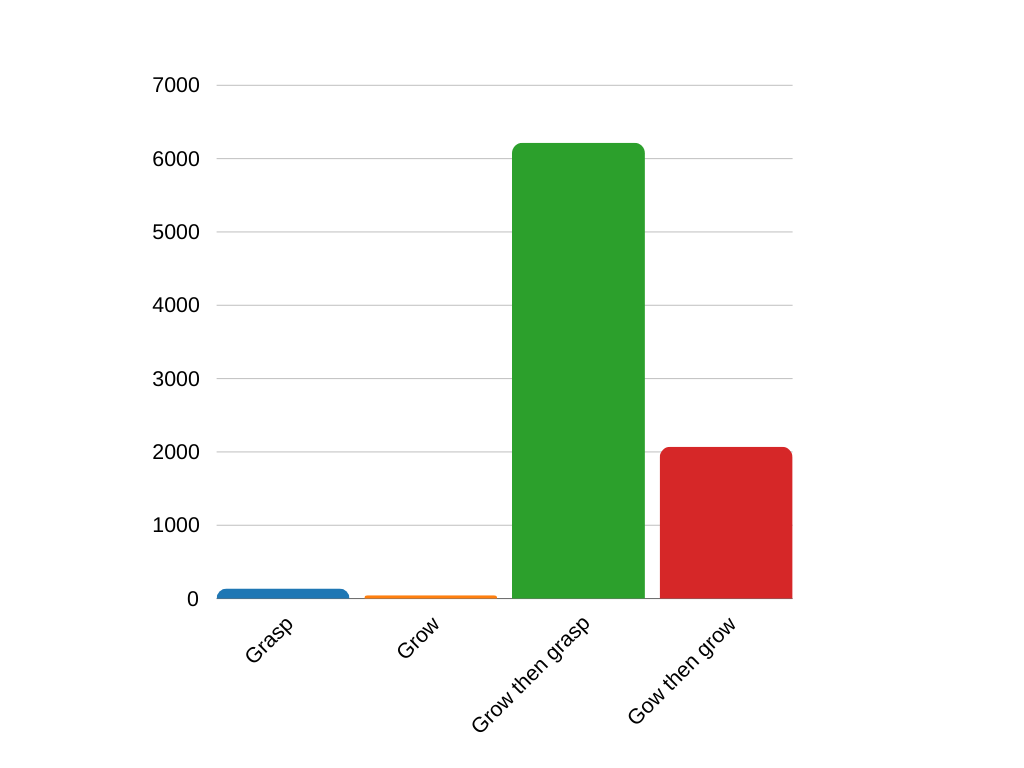}
    \caption{\textbf{Proportion of each goal type.} The distribution of goals is highly imbalanced, with sequential goals making up $98\%$ of the total.}
    \label{fig:goal_distribution}
\end{figure}

There are two types of goals in the environment: "\textit{Grasp \{object\}}" and "\textit{Grow \{object\}}." Additionally, we introduce sequential goals such as "\textit{Grow \{object A\} then grasp \{object B\}}" or "\textit{Grow \{object A\} then grow \{object B\}}." Rewards are set to 0 at each step, with a reward of 1 when achieving the goal. The agent is trained on goals sampled uniformly from all available goals. However, the distribution of goal types is not uniform, with the majority being sequential goals (Figure \ref{fig:goal_distribution}).

\section{Critic architecture}
\label{sec:archi}

In our setup, the critic is implemented as an MLP, positioned on top of the final decoder block. We experimented with different methods of integrating the critic with the actor:

\begin{enumerate}
    \item \textbf{Shared parameters with backpropagation:} The critic shares parameters with the actor and allows gradient backpropagation through these shared layers, enabling joint learning of the underlying representations.
    \item \textbf{Shared parameters without backpropagation:} In this case, the critic shares parameters with the actor, but gradients are backpropagated only through the critic's MLP head, keeping the shared layers frozen.
\end{enumerate}

We also explored two approaches for computing action Q-values:

\begin{enumerate}
    \item \textbf{Observation input:} The critic takes the observation as input and outputs Q-values for each possible action. This method is commonly used in discrete actions value-based algorithms like \citet{mnih2015human}.
    \item \textbf{Observation-action input:} The critic takes the concatenation of the observation and action as input, producing a single Q-value. This approach is more typical in continuous action spaces.
\end{enumerate}

\begin{figure}[htbp]
    \centering
    \includegraphics[width=\linewidth]{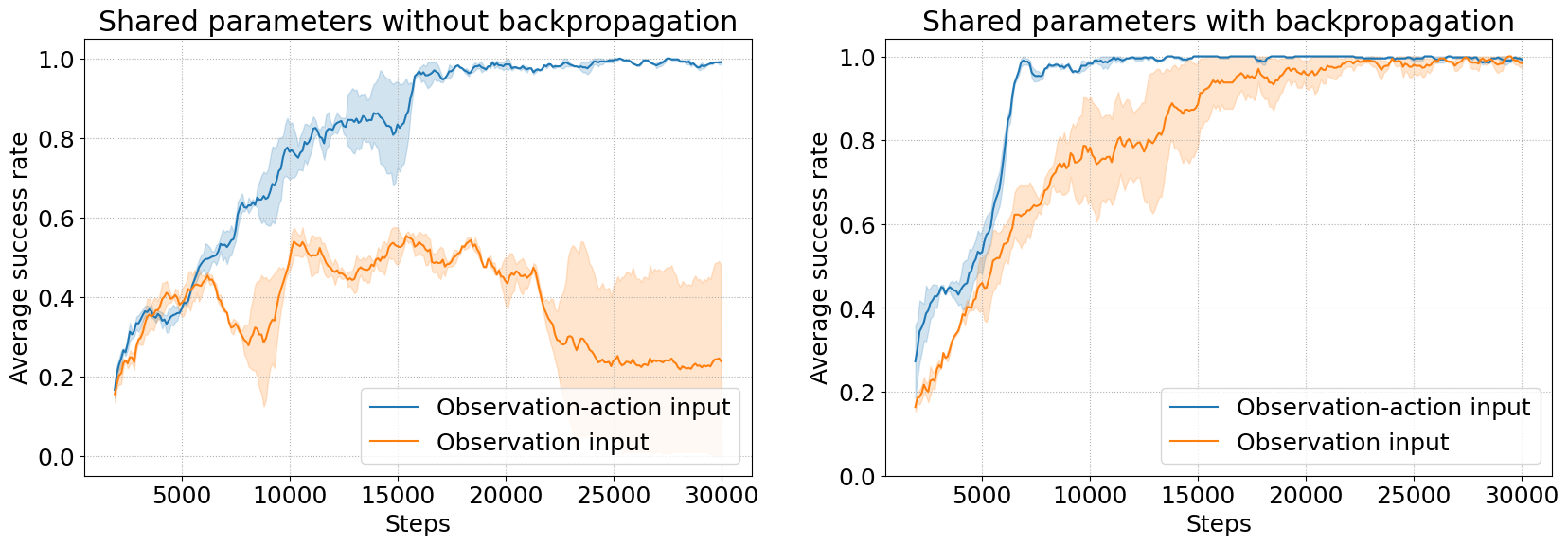}
    \caption{\textbf{Comparison of Critic Architectures.} The blue curve represents the architecture where both the observation and action are inputs, producing a single Q-value. The orange curve corresponds to the architecture where only the observation is input, producing Q-values for each action. The left plot illustrates the case where gradients are backpropagated only through the MLP head, while the right plot shows the case where gradients propagate through both the MLP and shared parameters. Mean and standard deviation are computed across two seeds.}
    \label{fig:sac_critic_architectures}
\end{figure}

These architecture variations were tested in a simplified version of the environment, which omits sequential goals and uses only \(N = 3\) objects (see Figure~\ref{fig:sac_critic_architectures}).

As shown in Figure~\ref{fig:sac_critic_architectures}, the critic architecture that uses both observation and action as inputs outperforms the one that relies solely on the observation. Additionally, the configuration where gradients are propagated through both the MLP and shared parameters yields better results compared to the version where gradients flow only through the MLP head.

\section{Implementation details}
\label{sec:implementation}

In our approach, the actor is modeled using a pre-trained encoder-decoder LLM, with the critic positioned as an MLP attached to the final decoder block, similar to the GLAM architecture. To optimize GPU VRAM usage during training, we employed LoRA \citep{hu_lora_2021} and 4-bit quantization techniques as described in \citep{dettmers_qlora_2023}.

We implemented SAC with automatic tuning of the entropy coefficient, initializing it at $0.005$ and setting a target entropy of $0$ for the SAC-GLAM version and initializing it at $0.05$ and setting a target entropy of $0.01$ in SAC-GLAM+HER, both using a learning rate of $2\mathrm{e}{-3}$. Both the actor and critic were trained using the Adam optimizer with a learning rate of $1\mathrm{e}{-4}$. Detailed hyperparameters are provided in Appendix~\ref{sec:hyperparameters}. The critic MLP has two hidden layers, each with 1024 units and ReLU activation functions.

We used a vectorized version of Playground-Text with 32 instances of the environment running (synchronously) in parallel. In order to accelerate the use of an LLM as the trained RL policy (and critic), we leveraged Lamorel\footnote{https://github.com/flowersteam/lamorel} as in GLAM and deployed 4 instances of the LLM in parallel (distributing both the forward passes to compute the actions' probability and the training in a Data Parallelism setting).
When using Flan-T5 250M, each LLM instance is distributed (Vertical Model Parallelism) over one Nvidia V100 32GB GPUs requiring thus a total of 4 Nvidia V100 32GB GPUs to run an experiment (1 GPU $\times$ 4 LLM instances). In total, each experiment and ablation requires 80 GPU hours per seed on the Nvidia V100 32GB.

\section{Update frequency and epochs} \label{sec:update_freq}
Update frequency usually plays a key role in Deep RL balancing between sample efficiency and overfitting to data collected. This issue is particularly pronounced in on-policy settings, where updates are based solely on the data gathered between each update. However, when considering very large policy models such as LLMs, properly setting the update frequency also becomes key in off-policy settings as updates may become slower than data collection. Moreover, both PPO and SAC possess a parameter controlling the number of gradient updates performed on collected data, which also affects sample efficiency. 

\begin{figure}[htbp]
    \centering
    \includegraphics[width=\linewidth]{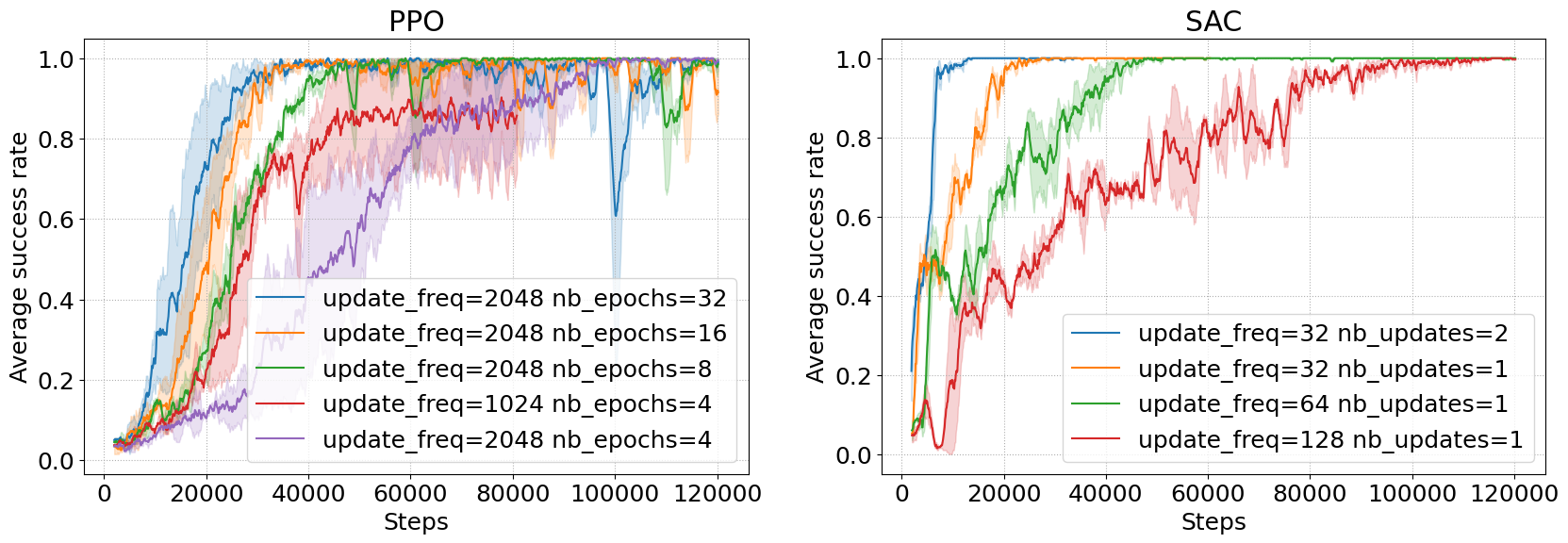}
    \caption{\textbf{Comparison of different update frequencies and epochs for PPO and SAC.} In PPO, \textit{nb\_epochs} refers to the number of times the model processes each transition during an update, while, for SAC, it represents the number of batches sampled from the replay buffer during each update.}
    \label{fig:update_freq}
\end{figure}

We studied different values for these parameters for both SAC and PPO on a simplified version of the environment. In the case of PPO, we observed that updating the model too frequently (every 1024 steps) leads to the method plateauing at a suboptimal policy. Additionally, performing too many epochs (32) at each update introduces instability in the training process (Figure \ref{fig:update_freq}). Concerning SAC, the method benefits from frequent updates with multiple epochs, inducing a balance between sample and time efficiency (frequent updates and multiple epochs imply slower training).

\section{Batch sampling with HER}
\label{sec:sampling}

\begin{figure}[htbp]
    \centering
    \includegraphics[width=0.7\linewidth]{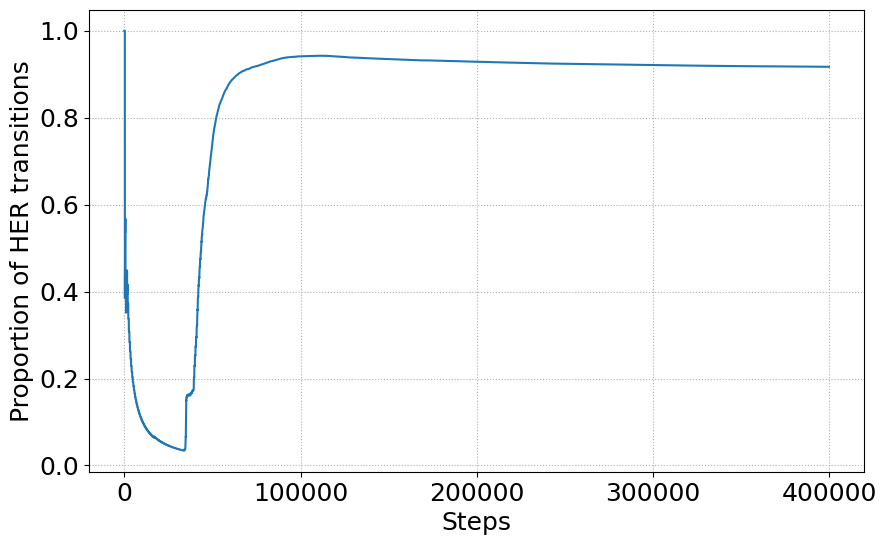}
    \caption{\textbf{Proportion of HER transitions in the replay buffer when uniformly sampling.} The plot shows the proportion of transitions relabeled by HER compared to those directly collected from the environment.}
    \label{fig:proportion_her}
\end{figure}

In order to optimize sample efficiency, we analyzed the impact of diversity in batches sampled from our replay buffer in order best exploit collected data. We use HER with the \textit{future} strategy proposed in \citet{andrychowicz2017hindsight}, using all the hindsight goals returned by the social partner, and found that the replay buffer was predominantly filled with relabeled trajectories, as shown in Figure \ref{fig:proportion_her}. 

This could present two potential issues: 
\begin{enumerate}

\item In our environment, "grasp" goals are subgoals of "grow" goals (i.e. achieving a "grow" goal requires the agent to first grasp an object). Similarly, "grow" goals are subgoals of more complex goals like "grow then grasp" or "grow then grow." This hierarchy means that once a goal is achieved, many of the trajectories added to the replay buffer by HER are from simpler goals, which should be a minority in this environment (Figure \ref{fig:goal_distribution}). 

\item Because HER adds only successful trajectories, the replay buffer becomes dominated by positive examples, with very few negative ones. However, negative trajectories are also important in the learning process.
\end{enumerate}

\begin{figure}[htbp]
    \centering
    \includegraphics[width=0.7\linewidth]{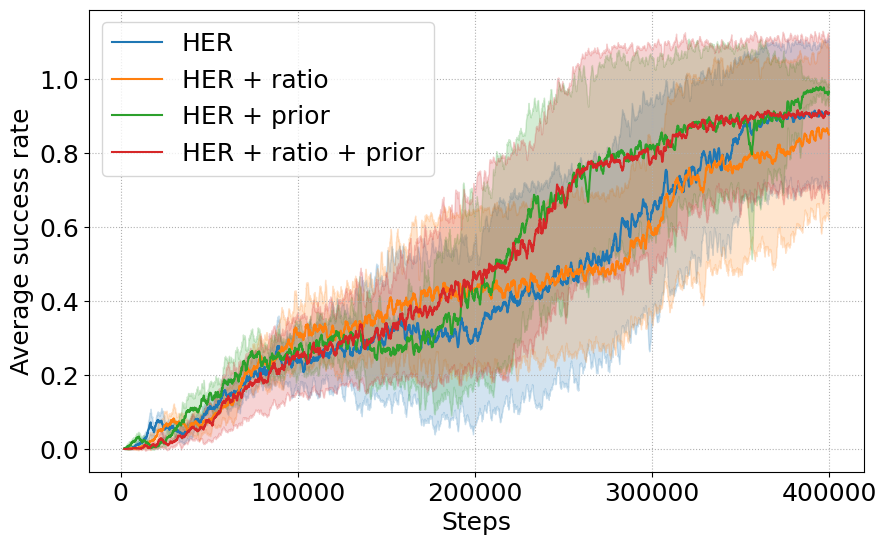}
    \caption{\textbf{Comparison of different sampling strategies.} "prior" refers to the sampling strategy used to address issue (1), while "ratio" refers to the strategy used to address issue (2).}
    \label{fig:her_ratio_sampling}
\end{figure}

To address issue (1), we adjusted the batch sampling strategy to align with the goal distribution of the environment. Instead of uniformly sampling from the replay buffer, batches were sampled with transitions that reflected the goal distribution illustrated in Figure \ref{fig:goal_distribution}. For issue (2), we balanced the batches by sampling 50\% of transitions from HER and 50\% from the environment. A similar approach was used in \citet{colas2020language}, where they also employed HER and sampled an equal proportion of positive and negative transitions, comparable to our method. We compared these methods in Figure \ref{fig:her_ratio_sampling}. These strategies did not affect sample efficiency significantly, so we opted to use the "HER + ratio" strategy in our experiments, as the "prior" method makes the strong assumption of knowing the environmental goal distribution.

\section{N-step returns parameters}
\label{sec:n_step}

\begin{figure}[htbp]
    \centering
    \includegraphics[width=\linewidth]{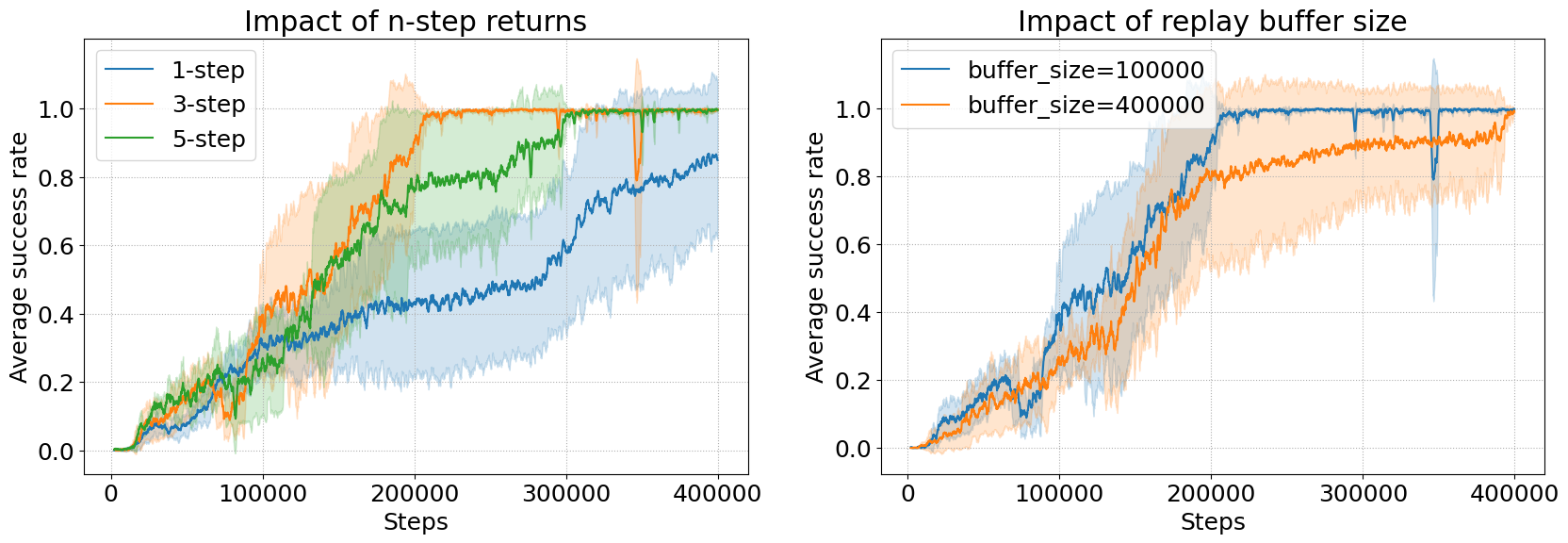}
    \caption{\textbf{Comparison of n-step parameters and replay buffer size.} The left plot illustrates the impact of the $n$ parameter (using a replay buffer with a capacity of $100000$), while the right plot shows the effect of the replay buffer capacity parameter (when using 3-step returns).}
    \label{fig:n_step}
\end{figure}

We looked at how the Temporal Difference loss was influencing our sample efficiency. Increasing the $n$ parameter of the n-step return increases the magnitude of the bootstrap, allowing returns to be propagated more quickly to earlier steps in the episode, which speeds up convergence. However, setting $n$ too high leads to highly off-policy updates, introducing instability. We tried to mitigate this instability by using a smaller replay buffer. We compared the performance of our method across different values of $n$ and various replay buffer sizes in Figure \ref{fig:n_step}. 

Based on these results, we chose to use 3-step returns in our experiments, as 5-step returns introduced instability. Additionally, we opted for a replay buffer with a capacity of $100000$, which provided greater stability compared to using a larger buffer.

\section{Hyperparameters}
\label{sec:hyperparameters}

Tables \ref{tab:ppo_param}, \ref{tab:sac_param} and  \ref{tab:sac_her_param} summarize all the hyperparameters used in the experiments whose results are presented in Figure~\ref{fig:sac_ppo}.

\begin{table}[htbp]
    \caption{PPO-GLAM hyperparameters}
    \centering 
    \begin{tabular}{ll}
    \hline
    \textbf{Variable} & \textbf{Value} \\
    \hline
    Number of transitions collected between two updates & $2048$ \\
    Number of epochs per update & $16$ \\
    Batch size & $256$ \\
    Entropy loss coefficient & $0.01$ \\
    Value function loss coefficient & $0.5$ \\
    Discount factor & $0.99$ \\
    Optimizer & Adam \\
    Learning rate & $1 \times 10^{-4}$ \\
    $\lambda$ factor of the Generalized Advantage Estimator & $0.99$ \\
    Clipping parameter $\epsilon$ & $0.2$ \\
    Maximum gradient norm & $0.5$ \\
    \hline
    \end{tabular}
    \label{tab:ppo_param}
\end{table}

\begin{table}[htbp]
    \caption{SAC-GLAM hyperparameters}
    \centering 
    \begin{tabular}{ll}
    \hline
    \textbf{Variable} & \textbf{Value} \\
    \hline
    Update frequency & $32$ \\
    Number of updates & $2$ \\
    Batch size & $256$ \\
    Discount factor & $0.99$ \\
    Optimizer & Adam \\
    Critic learning rate & $1 \times 10^{-4}$ \\
    Actor learning rate & $1 \times 10^{-4}$ \\
    Entropy coefficient & auto \\
    Entropy coefficient initialization & $0.005$ \\
    Target entropy & $0$ \\
    Entropy coefficient learning rate & $2 \times 10^{-3}$ \\
    n-step & $3$ \\
    Replay buffer capacity & $100000$ \\
    \hline
    \end{tabular}
    \label{tab:sac_param}
\end{table}

\begin{table}[htbp]
    \caption{SAC-GLAM + HER hyperparameters}
    \centering 
    \begin{tabular}{ll}
    \hline
    \textbf{Variable} & \textbf{Value} \\
    \hline
    Update frequency & $32$ \\
    Number of updates & $2$ \\
    Batch size & $256$ \\
    Discount factor & $0.99$ \\
    Optimizer & Adam \\
    Critic learning rate & $1 \times 10^{-4}$ \\
    Actor learning rate & $1 \times 10^{-4}$ \\
    Entropy coefficient & auto \\
    Entropy coefficient initialization & $0.05$ \\
    Target entropy & $0.01$ \\
    Entropy coefficient learning rate & $2 \times 10^{-3}$ \\
    n-step & $3$ \\
    Replay buffer capacity & $100000$ \\
    Hindsight proportion per batch & $0.5$ \\
    \hline
    \end{tabular}
    \label{tab:sac_her_param}
\end{table}

\end{document}